%% file: main.tex
\documentclass{article} 
\usepackage[preprint]{colm2026_conference}

\usepackage{microtype}
\usepackage{hyperref}
\usepackage{url}
\usepackage{booktabs}
\usepackage{multirow}
\usepackage{multicol}
\usepackage{graphicx}
\usepackage{amsmath}
\usepackage{makecell}
\usepackage{subcaption}
\usepackage{bbm}

\usepackage{lineno}

\definecolor{darkblue}{rgb}{0, 0, 0.5}
\hypersetup{colorlinks=true, citecolor=darkblue, linkcolor=darkblue, urlcolor=darkblue}

\title{Geometry-Aware CLIP Retrieval via Local Cross-Modal Alignment and Steering}

\begin{document}
\maketitle
\begin{center}

    {\Large
    \textbf{Nirmalendu Prakash}\textsuperscript{1,*}\quad
    \textbf{Narmeen Fatimah Oozeer}\textsuperscript{2,*}\quad
    \textbf{Xin Su}\textsuperscript{3}\quad
    \textbf{Phillip Howard}\textsuperscript{3}\quad
    \textbf{Shaan Shah}\textsuperscript{4}\quad
    \textbf{Zoe Wanying He}\textsuperscript{4}\quad
    \textbf{Shuang Wu}\textsuperscript{3}\quad
    \textbf{Shivam Raval}\textsuperscript{5}\quad
    \textbf{Roy Ka-Wei Lee}\textsuperscript{1}\quad
    \textbf{Meenakshi Khosla}\textsuperscript{4}\quad
    \textbf{Amir Abdullah}\textsuperscript{2,3}
    \par}

    \vspace{0.8em}

    {\Large
    \textsuperscript{1}Singapore University of Technology and Design \quad
    \textsuperscript{2}Martian \quad
    \textsuperscript{3}Thoughtworks \par
    \textsuperscript{4}UCSD \quad
    \textsuperscript{5}Harvard University
    }

    \vspace{0.5em}
    {\normalsize \textsuperscript{*}Equal contribution}
\end{center}

\begin{abstract}
CLIP retrieval is typically framed as a pointwise similarity problem in a shared embedding space. While CLIP achieves strong global cross-modal alignment, many retrieval failures arise from local geometric inconsistencies: nearby items are incorrectly ordered, leading to systematic confusions (e.g., pentagon vs. hexagon) and produces diffuse, weakly controlled result sets. Prior work largely optimizes for point wise relevance or finetuning to mitigate these problems.
We instead view retrieval as a problem of neighborhood alignment. Our work introduces (1) neighborhood-level re-ranking via Hungarian matching, which rewards structural consistency; (2) query-conditioned local steering, where directions derived from contrastive neighborhoods around the query reshape retrieval. We show that these techniques improve retrieval performance on attribute-binding and compositional retrieval tasks.
Together, these methods operate on local neighborhoods but serve different roles: re-ranking rewards alignment whereas local steering controls neighborhood structure. This shows that retrieval quality and controllability depend critically on local structure, which can be exploited at inference time without retraining.
\end{abstract}

\input{sections/introduction}
\input{sections/related_work}
\input{sections/dataset}

\input{sections/methodology}

\input{sections/metrics}

\input{sections/results}
\input{sections/Conclusion}




\bibliography{ref}
\bibliographystyle{colm2026_conference}
\clearpage
\appendix
\input{sections/appendix}

\end{document}

%% file: sections/introduction.tex
\section{Introduction}


Image search is a core capability in real-world systems, powering applications like e-commerce discovery~\citep{eclip}, video retrieval~\citep{videoclip} and even agriculture~\citep{agriclip}. It enables users to query large datasets using natural language or images, making retrieval more intuitive than keyword-based search. CLIP~\citep{radford2021learning} has revolutionized this space by enabling scalable, zero-shot cross-modal retrieval through a shared embedding space learned from large-scale image–text data.

However, real-world applications place diverse and often shifting demands on retrieval systems, requiring adaptation to new query types, domains, and notions of similarity. A single model must support fine-grained distinctions~\citep{finegrained}, compositional queries~\citep{gu2024language,zhong2024compositional}, and context-dependent interpretations across different data distributions.

A key challenge is that nearest-neighbor retrieval imposes a single global ranking over representations that may encode multiple valid relational structures. Polysemy implies that different semantic senses can occupy distinct directions or subspaces within an embedding~\citep{arora2018linear}. For example, the query ``bat'' may correspond to sports equipment, a superhero, or a flying animal, depending on context. As a result, there is often no single ``true'' neighborhood for a query, yet kNN retrieval must commit to one, leading to brittle or ambiguous rankings. In such cases, retrieval failures are local rather than global: the correct item is often present among nearby candidates but mis-ranked within the neighborhood.

A common response is to improve retrieval via fine-tuning or re-ranking~\citep{raclip,gao2024clip,rerank}, adapting the global representation or scoring function rather than modeling local cross-modal structure. Such methods primarily improve global alignment or query-conditioned scoring, rather than the structure of local neighborhoods themselves. Moreover, because they rely on contrastive objectives similar to CLIP, they favor globally consistent similarity; as shown in~\citep{wang2021understanding}, such objectives can distort or collapse finer-grained semantic structure, reducing sensitivity to context.

Rather than focusing solely on improving retrieval metrics such as Recall@k, we instead study the geometric mechanisms that govern retrieval behavior. We treat retrieval as a lookup of local neighborhood structure, and study how this structure is aligned across modalities. This leads to a more fundamental question: \textit{if retrieval is governed by local neighborhood structure, how can these neighborhoods be aligned across modalities without retraining?}

Recent work suggests that local neighborhood structure is more stable and semantically meaningful than global distances~\citep{groger2026revisiting}. Related findings in multimodal representation alignment further emphasize relational and compositional structure beyond pairwise distances~\citep{emnlp2025structure}, supporting our view of retrieval as the alignment of local cross-modal neighborhoods.

Building on this perspective, we make the following primary contributions: 
\begin{enumerate}
    \item \textbf{Local re-ranking as neighborhood alignment.} We frame retrieval as a local cross-modal alignment problem and show that many fine-grained retrieval failures arise from misordered local neighborhoods, and introduce a two-stage reranking procedure based on optimal transport that restores structural consistency among nearby candidates. This yields large gains in retrieval accuracy, improving CLIP R@1 from 42.8 to 78.4 on Urban1K.

    \item \textbf{Steering enables controllable attribute correction.} We show that query-conditioned concept directions can reshape the local retrieval neighborhood toward desired attributes, improving concept attribution and enabling fine-grained retrieval control without retraining.

    \item \textbf{Broad evaluation across structure-sensitive retrieval settings.} We evaluate on synthetic and real-image benchmarks spanning fine-grained, long-caption, and compositional retrieval, and show especially strong gains in multi-object compositional settings, where naive steering fails but structure-aware reranking improves R@10 from 0.230 to 0.682 on CLEVR-HOPE and R@1 from 0.091 to 0.725 on Visual Genome.
\end{enumerate}

\begin{figure}[t]  
    \centering
    \includegraphics[width=1\textwidth]{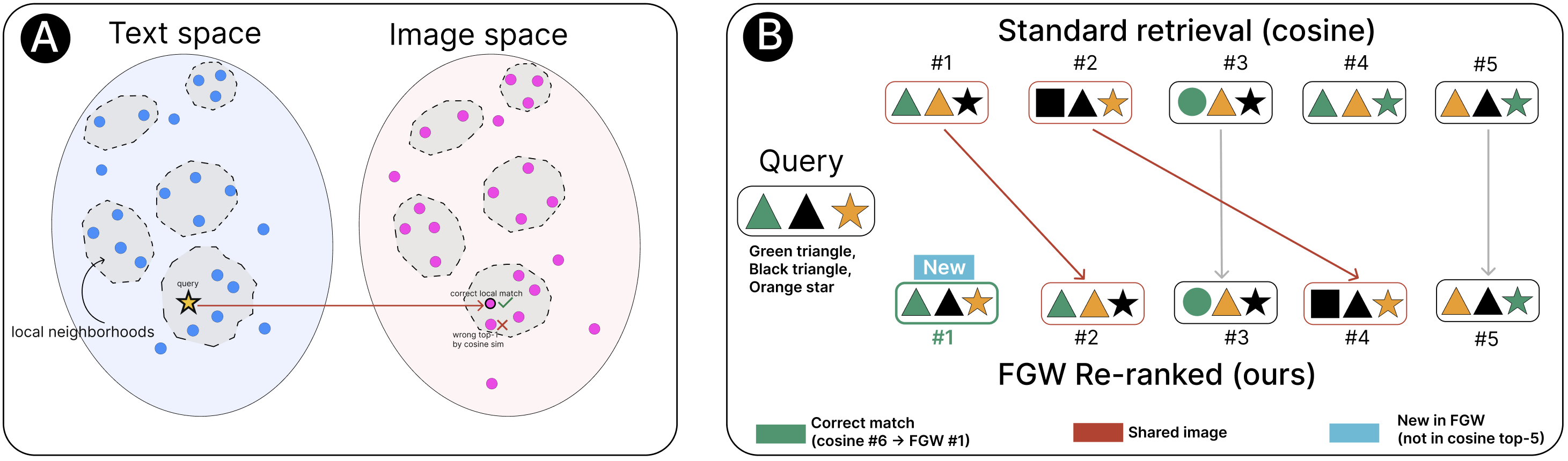}
    \caption{Local neighborhood structure provides information beyond pointwise similarity. The left panel illustrates how matching the geometry of nearby neighbors can identify the correct cross-modal correspondence when cosine similarity fails. The right panel shows an example from retrieval, where Fused Gromov Wasserstein(FGW)-based reranking exploits this local alignment and corrects the top-1 prediction.}
    \label{fig:introduction}
\end{figure}

%% file: sections/related_work.tex
\section{Related Work}

\paragraph{Improving cross-modal retrieval.}
A large body of work has sought to improve CLIP-style image-text retrieval by strengthening cross-modal alignment during training. Recent methods enhance retrieval using improved alignment objectives, such as soft-label supervision across and within modalities, or by introducing finer-grained matching between image and text tokens to better capture detailed correspondences \citep{huang2024cusa, asokan2025finelip}. Other approaches adapt retrieval at inference time through query-specific transformations of the embedding space, improving ranking without fully retraining the backbone \citep{xing2025quari}.

\paragraph{Controllable retrieval via query modification.}
Another line of work improves retrieval by modifying the input query itself. In particular, recent language-model-based methods enrich short or underspecified queries with additional visual detail, scene context, or quality-related attributes, enabling more controllable and semantically precise retrieval \citep{lu2026seeing}. These methods show that retrieval can be steered at inference time, but they typically rely on external generative models to rewrite or expand the query.

\paragraph{Representation geometry and local neighborhood structure.}
Our work is also motivated by recent studies of representational similarity across models and modalities. The Platonic Representation Hypothesis argues  sufficiently capable neural networks may converge toward shared representations \citep{huh2024platonic}. More recent work revisits this claim and shows that, after calibration, much of the agreement suggested by global similarity measures weakens substantially, while local neighborhood similarity remains comparatively robust across architectures and modalities \citep{groger2026aristotelian}. This suggests that local neighborhood relations may be a more stable cross-modal object than global similarity alone.

%% file: sections/dataset.tex
\section{Datasets}

We conduct our experiments using a variety of real-image and controlled benchmarks spanning long-caption retrieval, dense grounding, compositional reasoning, attribute binding, and symbolic shape matching. A brief description of each dataset is provided below;  see ~Appendix \ref{dataset_construction} for additional details.

\paragraph{Synthetic Shapes.}
Shapes are programmatically defined as SVG primitives: circles (\texttt{<circle>}), squares (\texttt{<rect>}), and polygons (triangle, pentagon, hexagon, star) via computed vertices. Primitives are placed at fixed positions on a white $224 \times 224$ canvas, rasterized with \texttt{cairosvg}, and converted to RGB images.

The dataset includes six shapes and seven colors (42 primitives). Each image contains three primitives sampled as unordered combinations with replacement, yielding 13{,}244 compositions. Captions are sorted, comma-separated color-shape pairs (e.g., ``blue circle, green star, red pentagon''). We also include 42 single-primitive reference images for per-concept analysis. This deterministic construction enables the study of compositional retrieval in a synthetic setting without the presence of background noise, yet surprisingly difficult for CLIP (which does not natively attain greater than 3.5\% accuracy on retrieval@1).

\paragraph{Urban1K.}
Urban1K \citep{zhang2024long} contains image-caption pairs of 
urban scenes, where each image is paired with a detailed long caption. We use it to test retrieval on real images with corresponding long queries.

\paragraph{Visual Genome.}
Visual Genome \citep{krishna2017visual} contains 
images with dense annotations, including region descriptions, objects, attributes, and relationships.
Its fine-grained object, attribute, and relation annotations make it suitable for detailed evaluation of cross-modal alignment.

\paragraph{Sugarcrepe.} SugarCrepe \citep{hsieh2023sugarcrepe} is a compositional benchmark in which each image is paired with a correct caption and a hard negative formed via a small semantic edit. For example, a swap-object negative may change ``a dog beside a bicycle'' to ``a bicycle beside a dog.'' This enables verification of compositional retrieval in a realistic setting.

\paragraph{Noun--Attributes--Contrastive (NAC).}
In order to study retrieval of real objects under different settings, we constructed an object-attribute dataset where attributes related to an object are changed while producing minimal modifications to the image background. For example, we generate images with prompts such as `upright bottle on a table' and `toppled bottle on a table', enabling precise measurement of CLIP's ability to differentiate attributes for a single object.



\paragraph{CLEVR-HOPE} CLEVR-HOPE \citep{berlot2024attribute} is a synthetic compositional retrieval benchmark built from CLEVR-style 3D scenes \cite{}. Each scene contains multiple objects annotated with four attributes: \emph{shape}, \emph{color}, \emph{material}, and \emph{size}. We use a subset of 1,000 images containing at least two objects, and construct queries directly from scene annotations (e.g.,
\emph{``a photo of a large red rubber sphere and a small blue metallic cube.''}). This dataset enables us to study compositions of steering vectors obtained for each of the individual attributes, such as `large sphere' and `blue cube'.

%% file: sections/methodology.tex
\section{Methodology}
\label{sec:methodology}
 
We study cross-modal retrieval: given a text query $q$ and a corpus of $N$ candidate images $\mathcal{C} = \{c_1, \ldots, c_N\}$, the goal is to rank candidates so that the most relevant images appear first. A dual-encoder model provides a text encoder $\psi$ and an image encoder $\phi$ that map inputs into a shared embedding space $\mathbb{R}^d$.
 
Our methodology follows a three-stage logical arc, where each stage motivates the next. First, standard cosine-similarity retrieval produces an initial candidate shortlist but fails on compositional scenes because it collapses multi-object structure into a single vector (Section~\ref{sec:bottleneck}). Second, we probe CLIP's embedding space using per-object text embeddings and find that precise local attribute directions exist but do not compose as they merge per-object representations into a single query vector destroying the structure needed for fine-grained retrieval (Section~\ref{sec:probing}). Third, motivated by this compositional failure, we introduce Fused Gromov--Wasserstein reranking, which takes the cosine shortlist and re-scores candidates by comparing query and image as structured sets of objects via optimal transport, preserving the per-object structure that merging destroys (Section~\ref{sec:fgw}).

\subsection{Preliminaries}
\label{sec:preliminaries}
 
\paragraph{Base retrieval.}
Given a text query $q$, we score each candidate by cosine similarity and rank by descending score:
\begin{equation}
\label{eq:cosine}
s(q, c_i) = \cos\bigl(\psi(q),\, \phi(c_i)\bigr), \qquad \pi = \operatorname{argsort}_i\bigl(-s(q, c_i)\bigr).
\end{equation}
 
\paragraph{Reranking.}
Let $S_k \subset \mathcal{C}$ denote the top-$k$ candidates from base retrieval. A reranker $r(q, c_i)$ re-scores this shortlist:
\begin{equation}
\label{eq:rerank}
\pi^* = \operatorname{argsort}_{c_i \in S_k}\bigl(-r(q, c_i)\bigr).
\end{equation}
Because $k \ll N$, the reranker may exploit richer structure than a single cosine score.

\subsection{The Structural Bottleneck}
\label{sec:bottleneck}
 
The core limitation of Eq.~\eqref{eq:cosine} is that $\phi(c_i)$ is a global embedding: it collapses all objects in an image into a single vector, discarding per-object information. Two candidates may contain the same objects in different spatial layouts yet project to nearly identical embeddings.
 
This is not merely a quantization issue. Our diagnostics reveal that the geometric structure of the two embedding spaces is only partially shared: the Pearson correlation between text-space and image-space pairwise distances is $r = 0.51$ across all pairs, with highly non-uniform agreement across categories (triangles: $r = 0.71$; circles: $r = -0.21$, meaning the within-shape color geometry is inverted). Post-hoc corrections such as a ridge regression mapper from text to image space reduce cross-modal feature distance by 85\%, but leave structural alignment essentially unchanged (Gromov--Wasserstein distance: $0.00580 \to 0.00578$). The mapper learns a global rotation, not a geometric correction.
 
Crucially, the problem is compositional: an optimal transport plan achieves a perfect identity mapping on single-item pairs, yet retrieval on compositional scenes fails systematically, for example, pentagon$\leftrightarrow$hexagon and star$\leftrightarrow$triangle confusions, because multiple concepts must share one vector. This motivates decomposing both query and candidate into per-object representations and aligning them structurally.
 
\paragraph{From one object to many.}
To see precisely what decomposition enables, consider the degenerate case: $Q = \{\psi(q)\}$ and $C_i = \{\phi(c_i)\}$, each containing a single vector. Any alignment-based score reduces to cosine similarity, because a single point has no internal geometry, there are no pairwise distances to preserve. This is exactly the regime where tied scores occur. Once we decompose into $M$ query objects and $N$ candidate objects, the intra-set distance matrices become non-trivial ($\binom{M}{2}$ pairwise distances per side) and a structural matching term can activate. The transition from $M = 1$ to $M > 1$ is therefore qualitative, not merely quantitative: it unlocks the structural channel entirely.

\subsection{Per-Object Text Embeddings and the Limits of Pointwise Composition}
\label{sec:probing}

Before introducing structural reranking, we define the per-object representation used throughout this work and ask a diagnostic question: can per-object embeddings be composed into a single query vector for retrieval, or does composition require structure-aware matching?

\paragraph{Per-object text embeddings.}
We construct per-object embeddings directly in CLIP text space from scene-level annotations, without using an object detector or region encoder. For an object with noun $n$ and attributes $\{a_1,\ldots,a_L\}$, we encode the composed phrase (e.g., ``large red rubber sphere'') with the CLIP text encoder $\psi$, using standard prompt templates and $\ell_2$ normalization:
\begin{equation}
\mathbf{e}(n,\{a_\ell\}) = \bar{\psi}\bigl(\text{``}a_1 \; a_2 \; \cdots \; a_L \; n\text{''}\bigr),
\end{equation}
where $\bar{\psi}$ denotes template-averaged, normalized text encoding. Because CLIP maps text and images into a shared embedding space, these per-object text vectors can be compared directly to image embeddings. We apply the same decomposition to both queries and candidates using their scene annotations.

\paragraph{Diagnostic.}
Per-object text embeddings are highly informative in isolation, but do not compose reliably into a single retrieval vector. On CLEVR-HOPE, individual object embeddings support near-perfect fine-grained attribute retrieval, yet merging multiple object vectors substantially degrades performance, dropping Recall@10 from 0.230 for the CLIP baseline to 0.094 under average merging. This shows that the relevant object-level information is present, but is lost when multiple objects are collapsed into a single vector. This motivates our structural reranking approach, which preserves and compares object-level structure instead of merging it.

Additional composition diagnostic results are provided in Appendix~\ref{app:compositional}.
\subsection{Structure-Aware Reranking via Fused Gromov--Wasserstein}
\label{sec:fgw}
 
Motivated by the failure of pointwise aggregation, we propose reranking retrieved candidates using Fused Gromov--Wasserstein (FGW) optimal transport, which compares query and candidate as structured sets of objects rather than single vectors.
 
\paragraph{Per-object decomposition.}
Let $Q = \{q_1, \ldots, q_M\}$ be per-object query embeddings and $C_i = \{c_i^1, \ldots, c_i^{N_i}\}$ be per-object candidate embeddings, both constructed using the text-based encoding described in Section~\ref{sec:probing}. This construction requires no object detector or region encoder; it operates entirely on scene annotations and CLIP's text encoder.
 
\paragraph{Cost matrices.}
We define a cross-set feature cost $D_{jl} = 1 - \cos(q_j, c_i^l)$, capturing dissimilarity between query object $j$ and candidate object $l$. We additionally define intra-set distance matrices $D^Q \in \mathbb{R}^{M \times M}$ (pairwise cosine distances among query objects) and $D^C \in \mathbb{R}^{N \times N}$ (pairwise cosine distances among candidate objects). These encode the internal geometry of each set---how objects relate to each other---independently of any cross-set comparison.
 
Together, these matrices define two axes of alignment: feature alignment (are corresponding objects similar, via $D$?) and structural alignment (do the two object sets have the same relational geometry, via $D^Q$ vs.\ $D^C$?). A reranker that considers only $D$---the linear assignment problem, solvable in $O(n^3)$ by the Hungarian algorithm---matches per-object features but cannot distinguish candidates whose objects match individually yet differ in relational configuration.
 
\paragraph{FGW objective.}
FGW optimal transport seeks a transport plan $T \in \mathbb{R}_{\geq 0}^{M \times N}$ that softly assigns query objects to candidate objects while jointly minimizing feature cost and structural distortion. The plan must satisfy marginal constraints: $\sum_l T_{jl} = \mu_j$ and $\sum_j T_{jl} = \nu_l$, where $\mu$ and $\nu$ are uniform distributions over query and candidate objects. The FGW reranking score is:
\begin{equation}
\label{eq:fgw}
r_{\mathrm{FGW}}(Q, C_i) = -\min_{T \in \Pi(\mu, \nu)} \Bigl[(1 - \beta)\langle D, T \rangle + \beta \cdot \mathrm{GW}(Q, C_i, T)\Bigr],
\end{equation}
where the two terms serve complementary roles. The Wasserstein term $\langle D, T \rangle = \sum_{jl} D_{jl}\, T_{jl}$ penalizes plans that assign dissimilar objects to each other. The Gromov--Wasserstein term compares the internal geometry of the two object sets:
\begin{equation}
\label{eq:gw}
\mathrm{GW}(Q, C_i, T) = \sum_{j, j', l, l'} L\bigl(D^Q_{jj'},\, D^C_{ll'}\bigr) \cdot T_{jl}\, T_{j'l'},
\end{equation}
where $L(a, b) = (a - b)^2$ is the squared loss. This term penalizes transport plans that distort pairwise relationships: if two query objects are far apart but their matched candidate objects are close together, the plan incurs a structural penalty. Crucially, GW compares only the pattern of within-set relationships, making it sensitive to relational geometry without requiring a shared feature space.
 
The parameter $\beta \in [0, 1]$ interpolates between two failure modes: at $\beta = 0$ the reranker matches objects correctly by features but accepts scrambled layouts; at $\beta = 1$ it preserves relational structure but ignores feature similarity. The composed objective requires both: the right objects in the right arrangement.
 
\paragraph{Hard vs.\ soft assignment and neighborhood ranking.}
Hungarian matching enforces a hard one-to-one assignment between query and candidate objects, yielding a single optimal-assignment cost per candidate. This is sufficient to identify a strong top-1 match---and indeed, Hungarian reranking produces large R@1 improvements across all datasets (Section~\ref{local_retrieval_results})---but it cannot produce a meaningful ranking over the full top-$k$ neighborhood. Because hard assignment maps each query object to exactly one candidate object, the resulting scores are coarsely quantized: many candidates receive identical or near-identical costs, creating ties that prevent differentiation beyond the top match. In contrast, FGW's soft transport allows fractional mass to distribute across multiple candidate objects, producing continuously graded scores that distinguish between candidates that are structurally close but not identical. This is the key reason FGW yields improvements across the full retrieval neighborhood (R@5, R@10, nDCG@K), whereas Hungarian is effective only at R@1.
 
\paragraph{Reranking procedure.}
For each candidate $c_i$ in the shortlist $S_k$: (1)~construct per-object query embeddings $Q$ and per-object candidate embeddings $C_i$ from scene annotations using the text encoding in Section~\ref{sec:probing}; (2)~build $D$, $D^Q$, $D^C$; (3)~solve the FGW optimization via conditional gradient (Frank--Wolfe), which linearizes the quadratic GW term at each iteration and solves a linear OT subproblem via Sinkhorn, typically converging in 10--50 iterations; (4)~use the negative optimal cost as the reranking score. The final ranking sorts candidates by descending $r_{\mathrm{FGW}}$. Because the solve runs only over $k$ candidates with small $M, N$ ($\leq 20$), the reranking step adds modest overhead.
 
\paragraph{Special cases and nesting.}
The full nesting is:
\begin{equation}
\label{eq:nesting}
\underbrace{\text{Cosine sim.}}_{M=1} \;\subset\; \underbrace{\text{Hungarian}}_{\beta=0,\; \text{hard}} \;\subset\; \underbrace{\text{Wasserstein}}_{\beta=0,\; \text{soft}} \;\subset\; \underbrace{\text{FGW}}_{\beta>0,\; \text{soft + structure}}.
\end{equation}
Each generalization adds a degree of freedom---decomposition, soft assignment, structural matching---while retaining the previous method as a special case. This nesting makes ablation clean: comparing $M = 1$ vs.\ $M > 1$ isolates decomposition; $\beta = 0$ vs.\ $\beta > 0$ isolates the GW term; hard vs.\ soft transport isolates fractional mass.

%% file: sections/metrics.tex
\section{Evaluation Metrics}

We evaluate retrieval quality using four complementary metrics. For CAS and nDCG, relevance is estimated using VLM-based scoring of the retrieved images; for Synthetic Shapes, where labels are deterministic and fully known, we instead use heuristic symbolic matching.

\begin{enumerate}
    \item \textbf{Recall@K (R@K).}
    Recall@K measures the fraction of queries for which at least one relevant image appears in the top-$K$ retrieved set:
    \[
    \mathrm{Recall@}K
    =
    \frac{1}{|Q|}
    \sum_{q \in Q}
    \mathbf{1}\!\left[
    \exists x \in \mathrm{Top}K(\tilde q) : x \text{ matches the target concept}
    \right].
    \]

    \item \textbf{Concept Attribution Score (CAS).}
    CAS measures how strongly the retrieved neighborhood reflects the target concept:
    \[
    A(c,\tilde q)
    =
    \frac{1}{k}
    \sum_{x \in N_k(\tilde q)} \mathrm{sim}(x,c),
    \]
    where \(\mathrm{sim}(x,c)\in[0,1]\) is a VLM-assigned confidence score indicating how strongly image \(x\) expresses concept \(c\). For Synthetic Shapes, this score is computed heuristically from exact symbolic overlap rather than VLM judgments.

    \item \textbf{Noun-level Concept Attribution Score (CAS-noun).}
    CAS-noun measures whether retrieved images contain the full query objects as grounded entities rather than only matching attributes in aggregate. Let \(\mathcal{O}(q)\) denote the set of query objects and \(\mathcal{T}(x)\) the set of grounded object tuples in retrieved image \(x\). We define
    \[
    \mathrm{CAS\mbox{-}noun}(x,q)
    =
    \frac{1}{|\mathcal{O}(q)|}
    \sum_{o \in \mathcal{O}(q)}
    \mathbf{1}[\,o \in \mathcal{T}(x)\,].
    \]
    Higher CAS-noun indicates better preservation of object--attribute binding.

    \item \textbf{nDCG@K.}
    nDCG@K measures graded ranking quality with respect to the full query:
    \[
    \mathrm{DCG@}K(q)
    =
    \sum_{i=1}^{K}\frac{r_i}{\log_2(i+1)},
    \qquad
    \mathrm{nDCG@}K(q)
    =
    \frac{\mathrm{DCG@}K(q)}{\mathrm{IDCG@}K(q)},
    \]
    where \(r_i \in \{0,1,2,3,4\}\) is a graded relevance score for the \(i\)-th retrieved image, assigned by a VLM for real-image datasets and by symbolic heuristics for Synthetic Shapes.
\end{enumerate}

Full definitions and implementation details are provided in Appendix~\ref{app:metrics}.

\section{Results}
\subsection{Locally Reranked Retrieval}
\label{local_retrieval_results}
\begin{table}[h]
\centering
\resizebox{\textwidth}{!}{%
\begin{tabular}{ll ccc ccc}
\toprule
 & & \multicolumn{3}{c}{Recall@$K$} & \multicolumn{3}{c}{nDCG@$K$} \\
\cmidrule(lr){3-5} \cmidrule(lr){6-8}
Stage 1 & Stage 2 & R@1 & R@5 & R@10 & nDCG@1 & nDCG@5 & nDCG@10 \\
\midrule
CLIP & None                &  3.5 & 13.4 & 21.3 & 0.4 & \textbf{0.6} & 0.7 \\
CLIP & Local Hungarian     & \textbf{16.4} & --- & --- & 0.2 & --- & --- \\
CLIP & Local FGW           & 5.1 & \textbf{14.9} & \textbf{22.5} & \textbf{0.5} & \textbf{0.6} & \textbf{0.8} \\
\midrule
Ridge & None               & 72.3 & 91.5 & 94.6 & 0.9 & \textbf{0.8} & \textbf{0.9} \\
Ridge & Local Hungarian    & \textbf{97.6} & --- & ---  & \textbf{1.0} & --- & --- \\
Ridge & Local FGW          & 81.3 & \textbf{94.1} & \textbf{95.7} & 0.4 & 0.7 & 0.7 \\
\midrule
Ridge+steer & None            & 72.8 & \textbf{100.0} & \textbf{100.0} & \textbf{0.7} & \textbf{0.8} & 0.8 \\
Ridge+steer & Local Hungarian & 97.6 & --- & --- & \textbf{1.0} & --- & --- \\
Ridge+steer & Local FGW       & \textbf{100.0} & \textbf{100.0} & \textbf{100.0} & 0.6 & 0.7 & \textbf{0.9} \\
\bottomrule
\end{tabular}%
}
\caption{Synthetic Shapes. Best per-metric in \textbf{bold} within each section. ``---'' = not applicable. Recall@$K$ from full retrieval; nDCG@$K$ from VLM-relevance aggregation ($n{=}200$ queries).}
\label{tab:results-synth}
\end{table}
 
\begin{table}[h]
\centering
\resizebox{\textwidth}{!}{%
\begin{tabular}{ll ccc ccc}
\toprule
 & & \multicolumn{3}{c}{Recall@$K$} & \multicolumn{3}{c}{nDCG@$K$} \\
\cmidrule(lr){3-5} \cmidrule(lr){6-8}
Stage 1 & Stage 2 & R@1 & R@5 & R@10 & nDCG@1 & nDCG@5 & nDCG@10 \\
\midrule
CLIP & None                & 42.8 & 67.3 & 75.9 & \textbf{0.7} & \textbf{0.7} & \textbf{0.8} \\
CLIP & Local Hungarian     & \textbf{78.4} & \textbf{83.3} & \textbf{86.1} & \textbf{0.8} & --- & --- \\
CLIP & Local FGW           & 56.3 & 72.7 & 79.3 & 0.5 & 0.6 & \textbf{0.8} \\
\midrule
Ridge & None               & 74.3 & 84.4 & 87.2 & \textbf{0.8} & \textbf{0.8} & \textbf{0.9} \\
Ridge & Local Hungarian    & \textbf{91.1} & --- & --- & --- & --- & --- \\
Ridge & Local FGW          & 83.6 & \textbf{88.0} & \textbf{89.5} & 0.4 & 0.6 & 0.8 \\
\midrule
\end{tabular}%
}
\caption{Urban1K. Best per-metric in \textbf{bold} within each section. ``---'' = not applicable. Recall@$K$ from full retrieval; nDCG@$K$ from VLM-relevance aggregation ($n{=}200$ queries).}
\label{tab:results-urban}
\end{table}

%% file: sections/results.tex
The results of the reranking methods are shown in Tables~\ref{tab:results-synth}, \ref{tab:results-urban}, and \ref{tab:results-sugarcrepe}. Across datasets, reranking yields substantial improvements in R@K over the cosine baseline. We also observe that Hungarian reranking outperforms local FGW in several settings. We discuss this difference briefly in Appendix~\ref{hungarian_vs_fgw}, and leave a more systematic analysis to future work. We additionally replicate these experiments with SigLIP~\citep{zhai2023sigmoid}, and present the corresponding results in Appendix~\ref{app:siglip_results}.
 


\subsection{Compositional Steering}

\input{sections/compositional_steering}

\subsection{Steering for Concept Control}

Beyond reranking, we explore \emph{steering for concept control} as a fine-grained retrieval setting. Using locally derived steering vectors, we bias retrieval toward desired attributes and away from contrasting ones, enabling targeted control over the retrieved set. This complements reranking: rather than improving order alone, it allows retrieval to be steered along interpretable semantic directions. We present qualitative and quantitative results in Appendix~\ref{app:steering_concept_control}, and leave a fuller analysis to future work.

%% file: sections/compositional_steering.tex
\begin{table*}[t]
\centering
\small
\begin{tabular}{llccccc}
\toprule
Dataset & Metric
& \makecell{Baseline\\CLIP}
& \makecell{Multi-object\\steering\\(average)}
& \makecell{FGW\\reranking}
& \makecell{Hungarian\\reranking} \\
\midrule
\multirow{8}{*}{CLEVR-HOPE}
& R@1         & 0.050 & 0.010 & 0.682 & \textbf{0.683} \\
& R@5         & 0.141 & 0.051 & 0.682 & \textbf{---} \\
& R@10        & 0.230 & 0.094 & 0.682 & \textbf{---} \\
& nDCG@1      & 0.054 & 0.012 & 0.683 & \textbf{0.695} \\
& nDCG@5      & 0.103 & 0.039 & \textbf{0.665} & --- \\
& nDCG@10     & 0.146 & 0.063 & 0.682 & --- \\
& CAS-gen@10  & 0.855 & 0.785 & 0.923 & --- \\
& CAS-noun@10 & 0.118 & 0.078 & 0.264 & --- \\
\midrule
\multirow{8}{*}{Visual Genome}
& R@1         & 0.091 & 0.068 & 0.249 & \textbf{0.725} \\
& R@5         & 0.199 & 0.148 & \textbf{0.399} & --- \\
& R@10        & 0.259 & 0.208 & \textbf{0.462} & --- \\
& nDCG@1      & 0.095 & 0.073 & 0.249 & \textbf{0.733} \\
& nDCG@5      & 0.155 & 0.118 & \textbf{0.327} & --- \\
& nDCG@10     & 0.178 & 0.141 & \textbf{0.351} & --- \\
& CAS-gen@10  & \textbf{0.398} & 0.389 & 0.361 & --- \\
& CAS-noun@10 & 0.057 & 0.049 & \textbf{0.078} & --- \\
\bottomrule
\end{tabular}
\caption{Compositional retrieval results on CLEVR-HOPE and Visual Genome. Queries are constructed from multiple object--attribute descriptions, and we compare baseline CLIP retrieval, naive multi-object steering, and FGW/Hungarian reranking. Reranking consistently improves retrieval and ranking metrics, and yields the strongest noun-level grounding (CAS-noun), while naive multi-object steering degrades performance in both settings.}
\label{tab:compositional_combined}
\end{table*}


We evaluate compositional retrieval on both a synthetic benchmark (\textbf{CLEVR-HOPE}) and a real-image benchmark (\textbf{Visual Genome}), where each query describes multiple attributed objects. The main challenge is correct object--attribute binding. As shown in Table~\ref{tab:compositional_combined}, naive multi-object steering consistently hurts performance relative to baseline CLIP, suggesting that merging independently steered object representations introduces interference. In contrast, \emph{FGW} reranking substantially improves retrieval and ranking metrics on both datasets, with especially large gains in R@10 and CAS-noun. On Visual Genome, baseline CLIP remains slightly stronger on CAS-gen@10, indicating that FGW favors more precise object-level grounding over aggregate attribute coverage. Overall, these results show that preserving object-level structure is more effective for compositional retrieval than collapsing multiple object updates into a single query vector.
Full analysis and metric definitions are deferred to Appendix~\ref{app:compositional_steering}.

%% file: sections/Conclusion.tex
\section{Conclusion}

We studied CLIP retrieval from the perspective of local cross-modal geometry.
Our main findings are as follows.

\begin{enumerate}
    \item \textbf{Retrieval errors are often local reordering failures.}
    The correct item is frequently present in the top-$k$ set but
    mis-ranked among nearby distractors.

    \item \textbf{Retrieval is locally controllable.} Query-conditioned
    perturbations alter nearby rankings while leaving distant items
    unchanged, confirming that retrieval behavior is governed by local
    structure.

    \item \textbf{Structural re-ranking improves fine-grained retrieval.}
    FGW reranking yields consistent gains in compositional and
    attribute-binding settings where pointwise similarity alone fails.
    Naive merging of per-object vectors introduces severe interference,
    while structure-aware matching remains effective by preserving
    object-level relations.

    \item \textbf{Neighborhood quality matters more than top-1 in
    practice.} Deployed systems return ranked sets, not single answers.
    FGW's largest gains appear at R@5, R@10, and nDCG@K because
    structural matching preserves relational geometry across the
    neighborhood, not just pointwise relevance at rank~1.
\end{enumerate}

\paragraph{Future work.}
Our methods exploit local structure post-hoc. A natural next step is to
incorporate neighborhood alignment directly into training, for example,
by penalizing cross-modal structural distortion within local
neighborhoods during pretraining. A complementary direction is to study
\emph{when} local alignment forms during training, revealing where
targeted intervention would be most effective and whether contrastive
objectives systematically neglect local structure in favor of global
alignment.

%% file: sections/appendix.tex
\section{Per-Object Text Embeddings and Composition Diagnostics}
\label{app:compositional}

We provide additional details on the per-object representation used for structural reranking, together with the diagnostic results showing why pointwise composition fails in multi-object retrieval.

\paragraph{Per-object text embeddings.}
Each object is represented from its scene annotation alone, without using an object detector or region encoder. For an object with noun $n$ and attributes $\{a_1,\ldots,a_L\}$, we form a descriptive phrase (e.g., ``large red rubber sphere'') and encode it with the CLIP text encoder $\psi$ using standard prompt templates, averaging over templates and $\ell_2$-normalizing:
\begin{equation}
\mathbf{e}(n,\{a_\ell\}) = \bar{\psi}\bigl(\text{``}a_1 \; a_2 \; \cdots \; a_L \; n\text{''}\bigr),
\end{equation}
where $\bar{\psi}$ denotes template-averaged, normalized text encoding.

We apply the same decomposition to both queries and candidates using their scene annotations: scene graphs for CLEVR-HOPE, object--attribute annotations for Visual Genome, and symbolic labels for Synthetic Shapes.

\paragraph{Example.}
Consider a scene with two objects:
\begin{equation*}
\texttt{object\;1:\;large\;red\;rubber\;sphere}, \qquad
\texttt{object\;2:\;small\;blue\;metallic\;cube}.
\end{equation*}
Standard CLIP retrieval encodes the full query as a single vector, whereas our decomposition produces
\begin{align}
\mathbf{e}_1 &= \bar{\psi}(\text{``large red rubber sphere''}), \\
\mathbf{e}_2 &= \bar{\psi}(\text{``small blue metallic cube''}).
\end{align}
Each candidate image is decomposed analogously, yielding a structured set of per-object vectors on both sides.

\paragraph{Composition diagnostic.}
We find that CLIP's text space contains meaningful per-object directions, but that these directions do not compose reliably into a single query vector. For individual objects, per-object retrieval achieves near-perfect fine-grained attribute matching on CLEVR-HOPE. However, when multiple per-object vectors are merged using averaging, min-pooling, or softmin aggregation, retrieval degrades substantially below the CLIP baseline. In particular, Recall@10 drops from 0.230 to 0.094 under average merging.

Interference analysis shows that 998 out of 1{,}000 queries degrade on at least one attribute slot, and the effect becomes universal for scenes with at least six objects. Color is the most disruptive attribute: when color improves, shape degrades in 81\% of cases and material in 80\%.

These results show that the information needed for fine-grained retrieval is present in per-object text embeddings, but collapsing multiple objects into a single vector destroys the compositional structure needed to use it effectively.
\section{Retrieval Results}
Refer Table \ref{tab:results-sugarcrepe} for retrieval results on SugarCrepe.
\begin{table}[h]
\centering

\resizebox{\textwidth}{!}{%
\begin{tabular}{ll ccc ccc}
\toprule
 & & \multicolumn{3}{c}{Recall@$K$} & \multicolumn{3}{c}{nDCG@$K$} \\
\cmidrule(lr){3-5} \cmidrule(lr){6-8}
Stage 1 & Stage 2 & R@1 & R@5 & R@10 & nDCG@1 & nDCG@5 & nDCG@10 \\
\midrule
CLIP & None                & 8.7 & 34.9 & 55.5 & \textbf{0.6} & \textbf{0.7} & \textbf{0.8} \\
CLIP & Local Hungarian     & 43.9 & --- & --- & 0.5 & --- & --- \\
CLIP & Local FGW           & 8.7 & 34.9 & 55.7 & 0.5 & 0.6 & 0.7 \\
\midrule
Ridge & None               & 9.2 & 39.9 & 62.8 & \textbf{0.6} & \textbf{0.7} & \textbf{0.8} \\
Ridge & Local Hungarian    & 52.5 & --- & --- & 0.5 & --- & --- \\
Ridge & Local FGW          & 11.5 & 42.2 & 64.7 & 0.4 & 0.5 & 0.7 \\
\midrule
Ridge+steer & None         & 9.28 & 39.9 & 62.8 & \textbf{0.6} & \textbf{0.7} & \textbf{0.8} \\
Ridge+steer & Local Hungarian & 55.0 & --- & --- & 0.5 & --- & --- \\
Ridge+steer & Local FGW    & \textbf{19.2} & \textbf{78.3} & \textbf{100.0} & 0.4 & 0.5 & 0.7 \\
\bottomrule
\end{tabular}%
}
\caption{SugarCrepe. Best per-metric in \textbf{bold}. ``---'' = not applicable. Recall@$K$ from full retrieval; nDCG@$K$ from VLM-relevance aggregation ($n{=}200$ queries).}
\label{tab:results-sugarcrepe}
\end{table}

\section{Local Retrieval Results on Siglip}
\label{app:siglip_results}
Results are shown in Tables 5–7.

\begin{table}[h]
\centering

\resizebox{\textwidth}{!}{%
\begin{tabular}{ll ccc ccc}
\toprule
 & & \multicolumn{3}{c}{Recall@$K$} & \multicolumn{3}{c}{nDCG@$K$} \\
\cmidrule(lr){3-5} \cmidrule(lr){6-8}
Stage 1 & Stage 2 & R@1 & R@5 & R@10 & nDCG@1 & nDCG@5 & nDCG@10 \\
\midrule
SigLIP & None                & 8.7 & 34.9 & 55.5 & \textbf{0.7} & \textbf{0.8} & \textbf{0.9} \\
SigLIP & Local Hungarian     & \textbf{43.9} & --- & --- & 0.4 & --- & --- \\
SigLIP & Local FGW           & 7.5 & 34.2 & 55.7 & 0.6 & 0.7 & 0.8 \\
\midrule
Ridge & None                 & \textbf{0.7} & 39.9 & 62.8 & \textbf{0.7} & \textbf{0.8} & 0.8 \\
Ridge & Local Hungarian      & 52.5 & --- & --- & 0.5 & --- & --- \\
Ridge & Local FGW            & 9.7 & \textbf{41.9} & \textbf{64.9} & 0.6 & 0.7 & 0.8 \\
\bottomrule
\end{tabular}%
}
\caption{SugarCrepe. Best per metric in \textbf{bold}. ``---'' = not applicable. Recall@$K$ from full retrieval; nDCG@$K$ from VLM-relevance aggregation ($n{=}200$ queries).}
\label{tab:results-sugarcrepe-siglip}
\end{table}

\begin{table}[h]
\centering

\resizebox{\textwidth}{!}{%
\begin{tabular}{ll ccc ccc}
\toprule
 & & \multicolumn{3}{c}{Recall@$K$} & \multicolumn{3}{c}{nDCG@$K$} \\
\cmidrule(lr){3-5} \cmidrule(lr){6-8}
Stage 1 & Stage 2 & R@1 & R@5 & R@10 & nDCG@1 & nDCG@5 & nDCG@10 \\
\midrule
SigLIP & None                & 3.5 & 13.4 & 21.3 & 0.4 & 0.6 & 0.7 \\
SigLIP & Local Hungarian     & 16.4 & --- & --- & 0.2 & --- & --- \\
SigLIP & Local FGW           & 4.5 & 16.1 & 23.9 & 0.5 & 0.6 & 0.8 \\
\midrule
Ridge & None                 & 72.3 & \textbf{91.5} & \textbf{94.6} & \textbf{0.9} & \textbf{0.8} & \textbf{0.9} \\
Ridge & Local Hungarian     & \textbf{97.6} & --- & --- & 1.0 & --- & --- \\
Ridge & Local FGW            & 18.3 & 84.0 & 93.4 & 0.5 & 0.7 & 0.8 \\
\bottomrule
\end{tabular}%
}
\caption{Synthetic shapes. Best per metric in \textbf{bold}. ``---'' = not applicable. nDCG@$K$ from VLM-relevance aggregation (full split, $n{=}13244$).}
\label{tab:results-synthetic-shapes-siglip}
\end{table}

\begin{table}[h]
\centering

\resizebox{\textwidth}{!}{%
\begin{tabular}{ll ccc ccc}
\toprule
 & & \multicolumn{3}{c}{Recall@$K$} & \multicolumn{3}{c}{nDCG@$K$} \\
\cmidrule(lr){3-5} \cmidrule(lr){6-8}
Stage 1 & Stage 2 & R@1 & R@5 & R@10 & nDCG@1 & nDCG@5 & nDCG@10 \\
\midrule
SigLIP & None                & 42.8 & 67.3 & 75.9 & 0.7 & 0.7 & 0.8 \\
SigLIP & Local Hungarian     & 76.5 & --- & --- & 0.8 & --- & --- \\
SigLIP & Local FGW           & 18.4 & 72.3 & 79.8 & 0.4 & 0.6 & 0.8 \\
\midrule
Ridge & None                 & 74.3 & \textbf{84.4} & 87.2 & \textbf{0.8} & \textbf{0.8} & \textbf{0.9} \\
Ridge & Local Hungarian     & \textbf{91.1} & --- & --- & 0.9 & --- & --- \\
Ridge & Local FGW            & 25.3 & 81.9 & \textbf{87.6} & 0.5 & 0.7 & 0.8 \\
\bottomrule
\end{tabular}%
}
\caption{Urban1k. Best per metric in \textbf{bold}. ``---'' = not applicable. nDCG@$K$ from VLM-relevance aggregation ($n{=}200$ queries).}
\label{tab:results-urban1k-siglip}
\end{table}

\section{Hungarian vs FGW Reranking}
\label{hungarian_vs_fgw}
\begin{figure}[h]  
    \centering
    \includegraphics[width=1\textwidth]{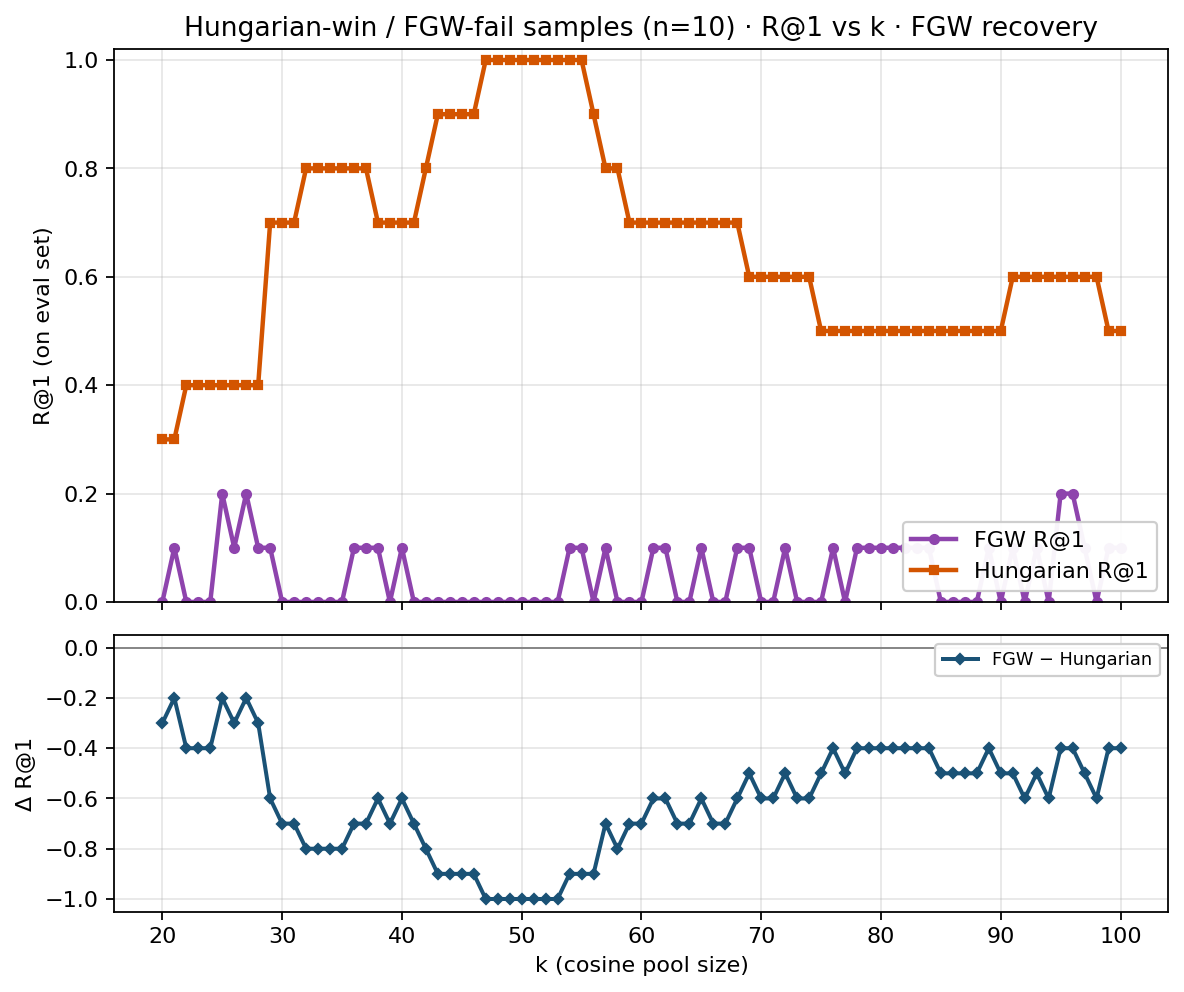}
    \caption{Synthetic Shapes failure analysis on queries where Hungarian achieves R@1 and FGW does not at the reference setting. We sweep the cosine candidate pool size \(k\) while keeping the FGW tradeoff parameter \(\alpha\) fixed. Top: fraction of queries in this subset for which each method attains R@1. Bottom: difference in R@1 between FGW and Hungarian. Hungarian remains consistently stronger across the sweep, while FGW recovers only sporadically, indicating that increasing \(k\) alone does not reliably fix the cases where structure-aware transport fails.}
    \label{fig:fgw_hungarian_recovery}
\end{figure}

\begin{figure}[h]
    \centering
    \begin{subfigure}[t]{0.49\textwidth}
        \centering
        \includegraphics[width=\textwidth]{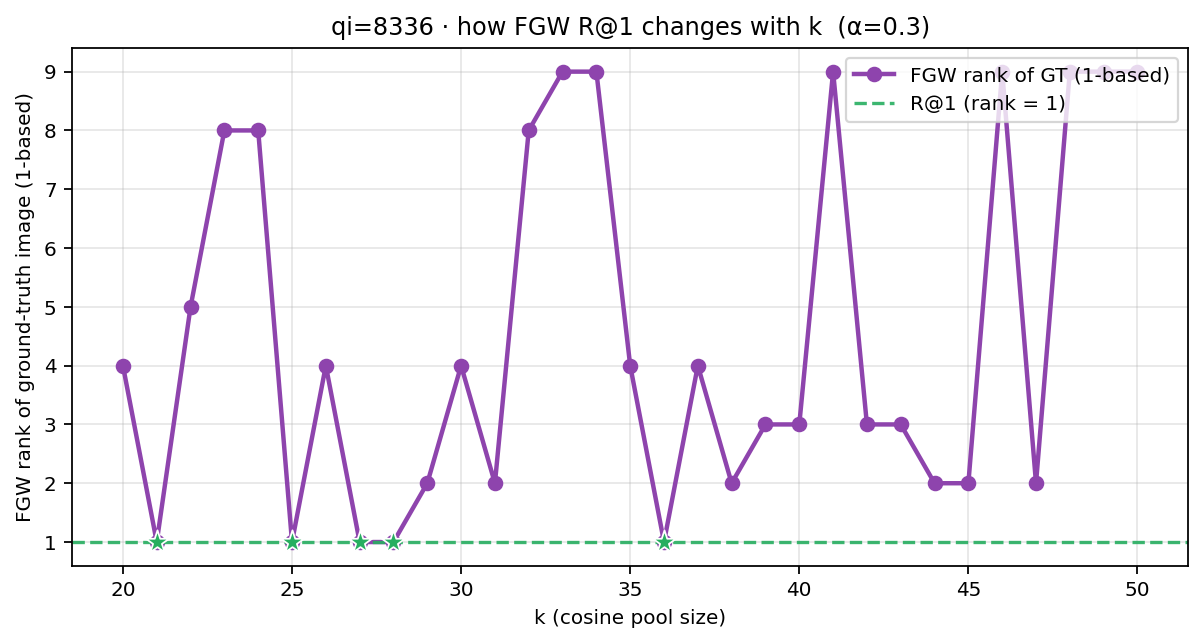}
        \caption{Query 8336.}
        \label{fig:fgw_rank_vs_k_q1}
    \end{subfigure}
    \hfill
    \begin{subfigure}[t]{0.49\textwidth}
        \centering
        \includegraphics[width=\textwidth]{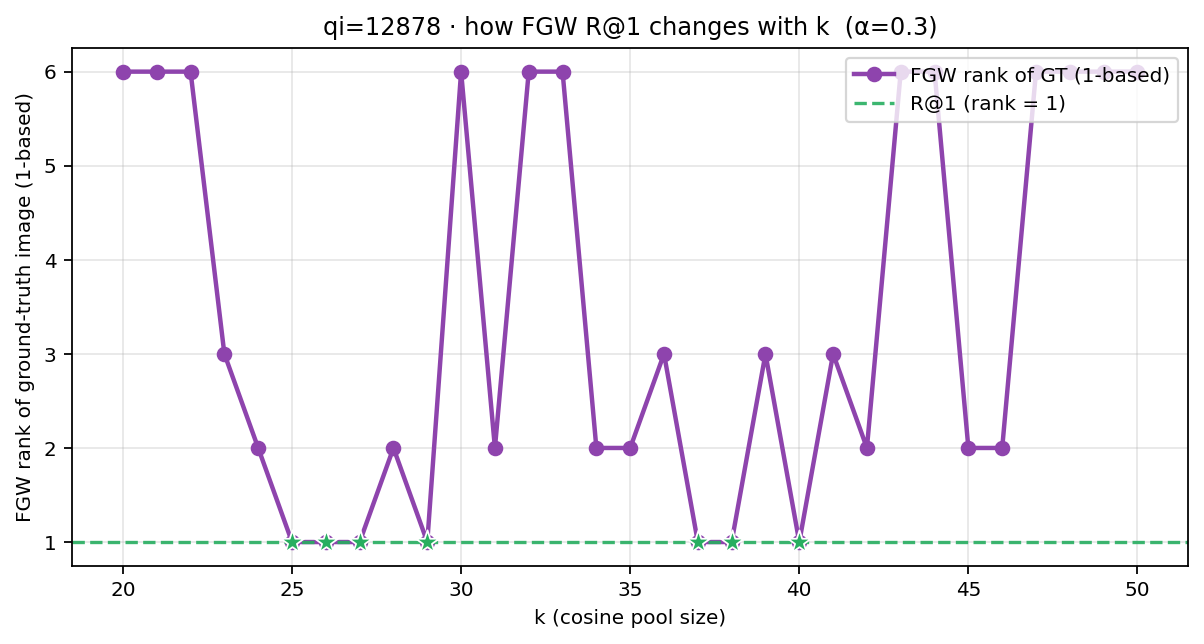}
        \caption{Query 12878.}
        \label{fig:fgw_rank_vs_k_q2}
    \end{subfigure}
    \caption{Representative per-query FGW recovery trajectories on Synthetic Shapes. For each query, we plot the rank of the ground-truth image under FGW as the cosine neighborhood size \(k\) varies, with the dashed line marking R@1. FGW recovery is non-monotonic: the ground-truth may briefly reach rank 0 for some values of \(k\), but this behavior is unstable and often disappears as the neighborhood changes.}
    \label{fig:fgw_rank_vs_k_examples}
\end{figure}

To understand why hard one-to-one matching can outperform structure-aware transport, we analyze Synthetic Shapes queries for which Hungarian reranking achieves R@1 while FGW does not. For this disagreement subset, we sweep only the cosine neighborhood size \(k\), while keeping the FGW tradeoff parameter \(\alpha\) fixed to the best value from the main FGW run. Figure~\ref{fig:fgw_hungarian_recovery} shows that Hungarian remains consistently strong across the entire \(k\)-range, whereas FGW recovers only intermittently and at much lower frequency. This suggests that, on these hard cases, the dominant signal is accurate one-to-one correspondence rather than the additional structural term used by FGW. In other words, enlarging the local neighborhood does not reliably resolve FGW's failures: recovery is possible for some queries, but it is unstable and does not close the gap to Hungarian.

\paragraph{Per-query recovery is unstable.}
Figure~\ref{fig:fgw_rank_vs_k_examples} shows representative per-query trajectories of the ground-truth rank under FGW as the cosine neighborhood size \(k\) varies. Recovery is highly non-monotonic: for some queries, the ground-truth reaches rank 0 only over a narrow range of \(k\), and then drops again as the neighborhood expands further. This indicates that FGW is sensitive to the precise local candidate composition, whereas Hungarian's discrete one-to-one assignment is more robust on these cases.

\section{Synthetic Shapes Failure Analysis}
\label{app:syn_shapes_confusion}
\begin{figure}[h]  
    \centering
    \includegraphics[width=0.7\textwidth]{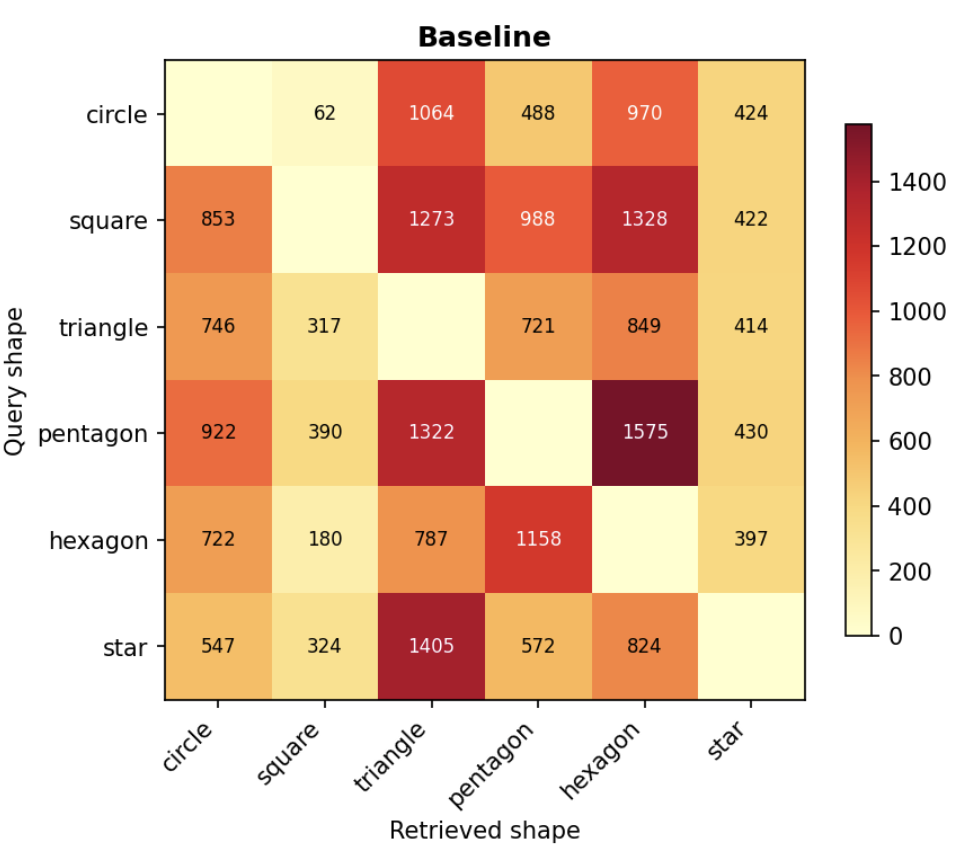}
    \caption{Shape substitution matrix on Synthetic Shapes. Each entry shows how often a query containing shape $s_i$ retrieves an image containing shape $s_j$ instead, with color held fixed. Errors are concentrated on a few structured confusions, most notably pentagon--hexagon.}
    \label{fig:syn_shape_confusion}
\end{figure}
To better understand baseline retrieval failures, we compute a \emph{shape substitution matrix} over the synthetic shapes dataset (Figure~\ref{fig:syn_shape_confusion}). Each entry records how often a query containing shape $s_i$ retrieves an image with shape $s_j$ instead, with color held fixed so that the matrix isolates shape-level confusion. The resulting matrix shows that CLIP does not fail uniformly: instead, errors are concentrated on a small number of recurring substitutions between visually similar shapes. In particular, pentagon--hexagon is the dominant failure mode, with additional confusions such as star--triangle and square--hexagon also appearing frequently. This indicates that CLIP's embedding space captures coarse semantic similarity but does not always preserve fine-grained local geometric distinctions. Motivated by this observation, we focus on pentagon--hexagon as a challenging case study and test whether targeted steering of the query embedding can reduce this structured confusion and improve retrieval.

\section{Dataset}
\label{tab:datasets}
Refer Table \ref{tab:dataset_summary} for the overall stats on the datasets used in this work.
\begin{table}[t]
\centering

\small
\setlength{\tabcolsep}{6pt}
\begin{tabular}{lccc}
\toprule
\textbf{Dataset} & \textbf{\#Samples} & \textbf{Domain} & \textbf{Setting} \\
\midrule
Synthetic Shapes & 2{,}430 & Synthetic & Compositional retrieval \\
Urban1K          & 1{,}000 & Real-world & Long-caption retrieval \\
Visual Genome    & 108{,}077  & Real-world & Dense grounding \\
SugarCrepe       & 7{,}512 & Real-world & Compositional retrieval \\
NAC              & 2{,}430 & Real-world (controlled) & Attribute binding \\
CLEVR-HOPE       & 1,000 & Synthetic & Compositional retrieval \\
\bottomrule
\end{tabular}
\caption{Summary of datasets used in our experiments.}
\label{tab:dataset_summary}
\end{table}

\subsection{Dataset Construction}
\label{dataset_construction}

\paragraph{Visual Genome.}
For each noun appearing in an image, we collect the deduplicated set of its annotated attribute strings, yielding a per-image mapping of the form \(\{\texttt{noun} \rightarrow [\texttt{attributes}]\}\).Overall, the corpus contains 65{,}557 unique attribute strings across 1{,}931{,}835 total attribute mentions.

We define a vocabulary of 80 contrastive attribute pairs grouped into 9 semantic categories: color (14 pairs), state (17), texture (9), shape (8), environment (8), size (6), pose (7), human (6), and material (5). Each contrastive pair is represented as two synonym sets rather than two single lexical items, e.g., \(\{\texttt{large}, \texttt{big}, \texttt{huge}, \texttt{giant}\}\) versus \(\{\texttt{small}, \texttt{little}, \texttt{tiny}, \texttt{miniature}\}\). This allows multiple surface forms to map to the same semantic pole and reduces sparsity from lexical variation.

For each contrastive pair, we identify the nouns grounded in the dataset for that pair. A noun \(n\) qualifies for a pair \((A^+, A^-)\) if it appears with at least one synonym from each pole in at least 10 images:
\[
\left|\left\{ i : n \in \mathrm{objects}(i),\ \mathrm{attrs}(n,i) \cap A^+ \neq \emptyset \right\}\right| \ge 10
\]
\[
\left|\left\{ i : n \in \mathrm{objects}(i),\ \mathrm{attrs}(n,i) \cap A^- \neq \emptyset \right\}\right| \ge 10
\]
Using this criterion, 76 of the 80 defined pairs have at least one qualifying noun. We then augment each image with a \texttt{contrastive\_attributes} field containing all instantiated \((\texttt{attr}_1,\texttt{attr}_2,\texttt{nouns})\) triples for which the image contains a qualifying noun associated with either side of the contrastive pair. The stats are shown in Table \ref{tab:vg_attr_distribution}.

For local steering, each example is constructed from a grounded noun--contrast pair by forming a source query \(\texttt{attr}_1\ \texttt{noun}\) and a target query \(\texttt{attr}_2\ \texttt{noun}\), and defining the steering direction as the normalized text-space difference between these two prompts. Because each contrastive pole is represented by a synonym set, evaluation is synonym-aware: for a query targeting attribute \(\texttt{attr}_2\), the positive set includes all corpus images in which the same noun appears with any synonym belonging to the target pole. Retrieval is counted as correct at rank \(k\) if at least one of these synonym-matched positives appears in the top-\(k\) results. This prevents the metric from penalizing lexical variation and ensures that performance reflects semantic, rather than purely string-level, alignment.

\begin{table}[t]
\centering
\small
\begin{tabular}{lrr}
\toprule
\textbf{Type} & \textbf{Count (both dirs)} & \textbf{\%} \\
\midrule
color             & 89,658  & 87.6 \\
size              & 4,500   & 4.4  \\
pose              & 1,788   & 1.7  \\
state             & 1,662   & 1.6  \\
shape             & 668     & 0.7  \\
human             & 216     & 0.2  \\
texture           & 72      & 0.1  \\
\midrule
TOTAL             & 102,318 & --   \\
\bottomrule
\end{tabular}
\caption{Distribution of contrastive attribute types in the Visual Genome-derived dataset. Counts are aggregated across both directions of each contrastive pair.}
\label{tab:vg_attr_distribution}
\end{table}


\paragraph{Noun-Attributes-Contrastive (NAC).}
\begin{itemize}
    \item \textbf{Attribute and noun curation.} We start from MIT-States nouns and adjective contrasts, and construct candidate binary attribute pairs from the adjective inventory and antonym lists. We further refine this pool using WordNet-based similarity scores together with manual filtering to remove near-duplicate, ambiguous, lighting-dependent, or overly generic attributes, as well as redundant nouns. This yields a curated set of noun-specific binary contrasts such as \textit{open} vs.\ \textit{closed} and \textit{wrinkled} vs.\ \textit{pressed}.
    
    \item \textbf{Prompt construction.} For each retained $(\text{noun}, \text{attr}_a, \text{attr}_b)$ tuple, we construct two prompts that share the same noun identity and scene context but differ only in the target attribute clause. Prompts are built from a noun-specific scene template when available, or otherwise a generic realistic scene description, together with an explicit clause describing the intended object or scene state. We additionally append generic realism constraints such as ``realistic details'' and ``no text, no watermark'' to reduce generation artifacts.
    
    \item \textbf{Contrastive image generation.} We generate each pair using the FLUX text-to-image model (\texttt{black-forest-labs/FLUX.1-dev}). The two images in a pair are sampled independently from scratch in a single batched call, using the same random seed but different attribute-specific prompts. This makes the pair contrastive: both images depict the same noun in matched scene context, while differing primarily in the intended binary attribute, without relying on editing or inpainting shortcuts.
    
    \item \textbf{Verification with language and vision-language models.} We apply a multi-stage verification pipeline to ensure that generated images exhibit the intended attributes and to identify additional plausible noun--attribute associations. First, we use GPT-4o-mini to predict which attributes are semantically plausible for each noun, producing a noun--attribute candidate list. Then, for each generated image, we use Qwen2.5-VL-32B-Instruct to perform per-attribute binary verification over applicable attributes. Specifically, given an image and noun, we query: \emph{``This is an image of a \{noun\}. Does the \{noun\} in this image exhibit the attribute `\{attr\}'? Answer with just Yes or No.''} We also request a confidence score in $[0,1]$, and retain only high-confidence positives. This open-vocabulary verification is applied not only to the prompted attribute, but also to other plausible attributes, enabling us to discover which noun--attribute pairs are reliably realized by the generator.
    
    \item \textbf{Filtering for steerability.} For steering experiments, we further restrict the dataset to noun--attribute pairs that are both visually verifiable and sufficiently frequent. Concretely, we retain only noun--attribute detections with Qwen confidence at least $0.9$, and require each noun--attribute pair to contain at least 10 verified samples so that retrieval metrics such as R@10 are well-defined. This produces a final set of steerable attribute contrasts over nouns for which the intended visual change is both meaningful and consistently present across multiple generated samples.
\end{itemize}

The final stats are shown in Table \ref{tab:nac_stats}.
\begin{table}[t]
\centering
\small
\begin{tabular}{ll}
\toprule
\textbf{Attribute contrast} & \textbf{Nouns retained (min count)} \\
\midrule
cluttered $\rightarrow$ tidy & garage (41), basement (15), kitchen (13) \\
foggy $\rightarrow$ clear & lake (19) \\
folded $\rightarrow$ unfolded & shirt (23) \\
frozen $\rightarrow$ thawed & lake (13) \\
open $\rightarrow$ closed & bucket (54), pot (41), bottle (27), bag (22), \\
& gate (20), road (20), street (20), door (19), \\
& garage (18), shower (16), window (15), town (14), \\
& box (12), book (11), library (10) \\
peeled $\rightarrow$ unpeeled & banana (17) \\
shattered $\rightarrow$ intact & glass (15), bowl (10) \\
sunny $\rightarrow$ overcast & lake (18) \\
toppled $\rightarrow$ upright & bottle (16), bucket (16), house (10) \\
\bottomrule
\end{tabular}
\caption{Final steerable attribute contrasts after verification and filtering. We retain only noun--attribute pairs with Qwen-VL confidence $\geq 0.9$ and at least 10 verified samples, ensuring that retrieval metrics such as R@10 can be computed reliably.}
\label{tab:nac_stats}
\end{table}

\section{VLM Evaluation}
\paragraph{Prompt for Query Alignment Score (QAS).}
To evaluate how well each retrieved image matches the \emph{full} query, we use the following prompt with a VLM, applied independently to each retrieved image:

\begin{quote}
\small
\textbf{You are evaluating image-to-text retrieval relevance.}

\textbf{Task:} Judge how well the image matches the full query.

\textbf{Instructions:}
\begin{itemize}
    \item Consider the full meaning of the query, not just one word.
    \item Use only visible evidence in the image.
    \item Consider objects, attributes, actions, relations, and scene details when relevant.
    \item Do not assume unstated details are present.
    \item If the image matches only part of the query, give a partial score.
    \item Return strict JSON only.
\end{itemize}

\textbf{Query:} ``\{query\}''

\textbf{Use this relevance scale:}
\begin{itemize}
    \item 0 = irrelevant
    \item 1 = weakly related
    \item 2 = partially matches
    \item 3 = mostly matches
    \item 4 = highly relevant and precise match
\end{itemize}

\textbf{Return JSON:}
\begin{verbatim}
{
  "relevance_score": 0-4,
  "confidence": float,
  "matched_elements": [...],
  "reason": str
}
\end{verbatim}
\end{quote}     

\paragraph{Attribute--Noun CAS Prompt.}
\begin{quote}\small
\textbf{You are evaluating whether an image contains a specific object with a specific attribute.}

\textbf{Task:}  
Decide whether the target object (noun) is present in the image \emph{and} has the target attribute.

\textbf{Instructions:}
\begin{itemize}
    \item Both conditions must be met: the noun must be visible and it must have the attribute.
    \item Do not count it as a match if the attribute applies to a different object.
    \item Base your answer only on visible evidence in the image.
    \item If the attribute--noun combination is only weakly supported or ambiguous, lower the confidence.
    \item Return strict JSON only.
\end{itemize}

\textbf{Target attribute:} \texttt{\{attribute\}} \\
\textbf{Target noun:} \texttt{\{noun\}}

\textbf{Return JSON with this schema:}
\begin{verbatim}
{
  "applicable": true or false,
  "confidence": float between 0 and 1,
  "reason": "short explanation based only on visible evidence"
}
\end{verbatim}
\end{quote}

\section{Metrics}
\label{app:metrics}
\paragraph{Recall@K (R@K).}
Recall@K measures the fraction of queries for which at least one of the top-$K$ retrieved images matches the target attribute. For a set of queries \(Q\), target concept \(c\), and steered query embedding \(\tilde{q}\), we define
\[
\mathrm{Recall@}K
=
\frac{1}{|Q|}
\sum_{q \in Q}
\mathbf{1}
\Big[
\exists x \in \mathrm{Top}K(\tilde{q}) : x \text{ has attribute } c
\Big],
\]
where \(\mathrm{Top}K(\tilde{q})\) denotes the top-$K$ retrieved images under cosine similarity to the steered query embedding \(\tilde{q}\). Higher Recall@K indicates that steering successfully brings at least one relevant image into the top retrieved set.

\paragraph{Concept Attribution Score (CAS).}
CAS measures how strongly the retrieved neighborhood reflects the target concept after steering. Given target concept \(c\), query embedding \(q\), and retrieved neighborhood \(N_k(q)\), we compute
\[
A(c, q)
=
\frac{1}{k}
\sum_{x \in N_k(q)} \mathrm{sim}(x, c),
\]
where \(\mathrm{sim}(x,c) \in [0,1]\) is the confidence score assigned by a VLM indicating whether image \(x\) exhibits concept \(c\). In practice, after steering the query toward concept \(c\), we retrieve the top-$k$ nearest images and average the VLM confidence over them. A higher CAS indicates that steering more consistently pulls retrieved images toward the intended concept.

\paragraph{Normalized Discounted Cumulative Gain (nDCG)@K.}
nDCG measures the graded relevance of the retrieved ranking with respect to the full query, while giving higher weight to items appearing earlier in the ranked list. For each query \(q\) with top-$K$ retrieved images \(\{x_1,\dots,x_K\}\), each image \(x_i\) is assigned a graded relevance score
\[
r_i = s(x_i, q) \in \{0,1,2,3,4\},
\]
where higher values indicate a stronger semantic match to the query. In the VLM-based setting, these scores are assigned by a vision-language model, with \(4\) denoting a precise match, \(3\) a strong match, \(2\) a partial match, \(1\) a weak match, and \(0\) an irrelevant image.

We first compute the discounted cumulative gain:
\[
\mathrm{DCG@}K(q)
=
\sum_{i=1}^{K}
\frac{r_i}{\log_2(i+1)}.
\]

To normalize for the best possible ordering of the same retrieved items, we compute the ideal discounted cumulative gain:
\[
\mathrm{IDCG@}K(q)
=
\sum_{i=1}^{K}
\frac{r_i^{\downarrow}}{\log_2(i+1)},
\]
where \(r_1^{\downarrow}, \dots, r_K^{\downarrow}\) are the same relevance scores sorted in descending order.

The normalized score is then
\[
\mathrm{nDCG@}K(q)
=
\frac{\mathrm{DCG@}K(q)}{\mathrm{IDCG@}K(q)}
\in [0,1].
\]

Aggregating over all queries gives
\[
\mathrm{Mean\mbox{-}nDCG@}K
=
\frac{1}{|Q|}
\sum_{q \in Q} \mathrm{nDCG@}K(q).
\]

Higher nDCG indicates that the retrieved set is not only relevant to the original query, but also correctly ordered so that the most relevant images appear earlier in the ranking.

\paragraph{Synthetic Shapes caveat.}
For the Synthetic Shapes dataset, relevance scores are not produced by a VLM. Instead, because each image has a deterministic symbolic label, we compute relevance heuristically from exact overlap between the target query and retrieved image labels. Concretely, both query and image are parsed into multisets of \((\texttt{color}, \texttt{shape})\) tuples, and the overlap count is scaled to the same \(0\)--\(4\) relevance range:
\[
r(x,q)
=
\mathrm{round}\!\left(
\frac{\mathrm{overlap}(x,q)}{\#\text{ target objects}} \cdot 4
\right).
\]
These heuristic scores are then used in the same nDCG@K computation as above. Thus, for Synthetic Shapes, nDCG reflects symbolic compositional overlap rather than VLM-judged semantic relevance.


\section{Steering for Concept Control}
\begin{figure}[h]  
    \centering
    \includegraphics[width=1\textwidth]{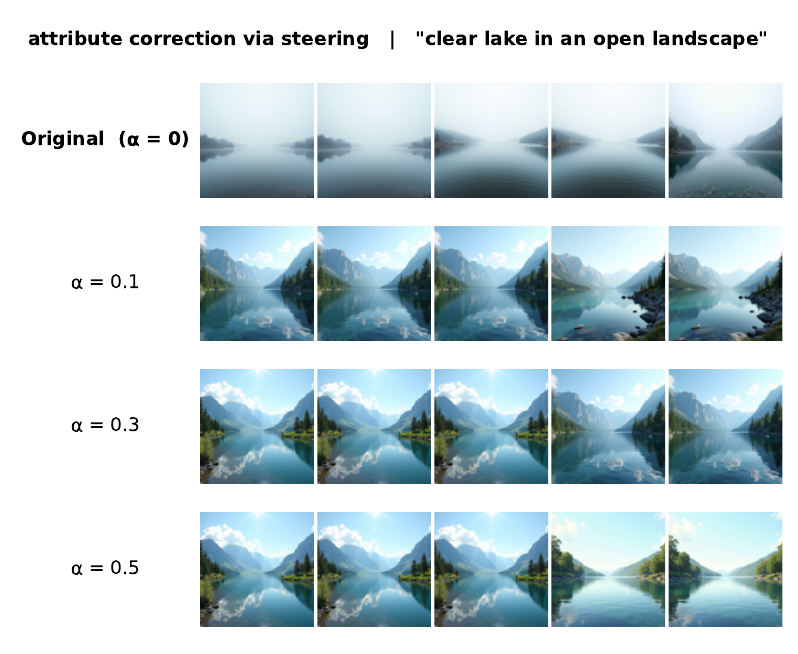}
    \caption{Sample NAC top-5 retrievals: original CLIP text-image matching ($\alpha = 0$) versus text embeddings steered to correct the attribute `clear' ($\alpha > 0$).}
    \label{fig:fine_steering_sample}
\end{figure}

\begin{figure}[h]  
    \centering
    \includegraphics[width=1\textwidth]{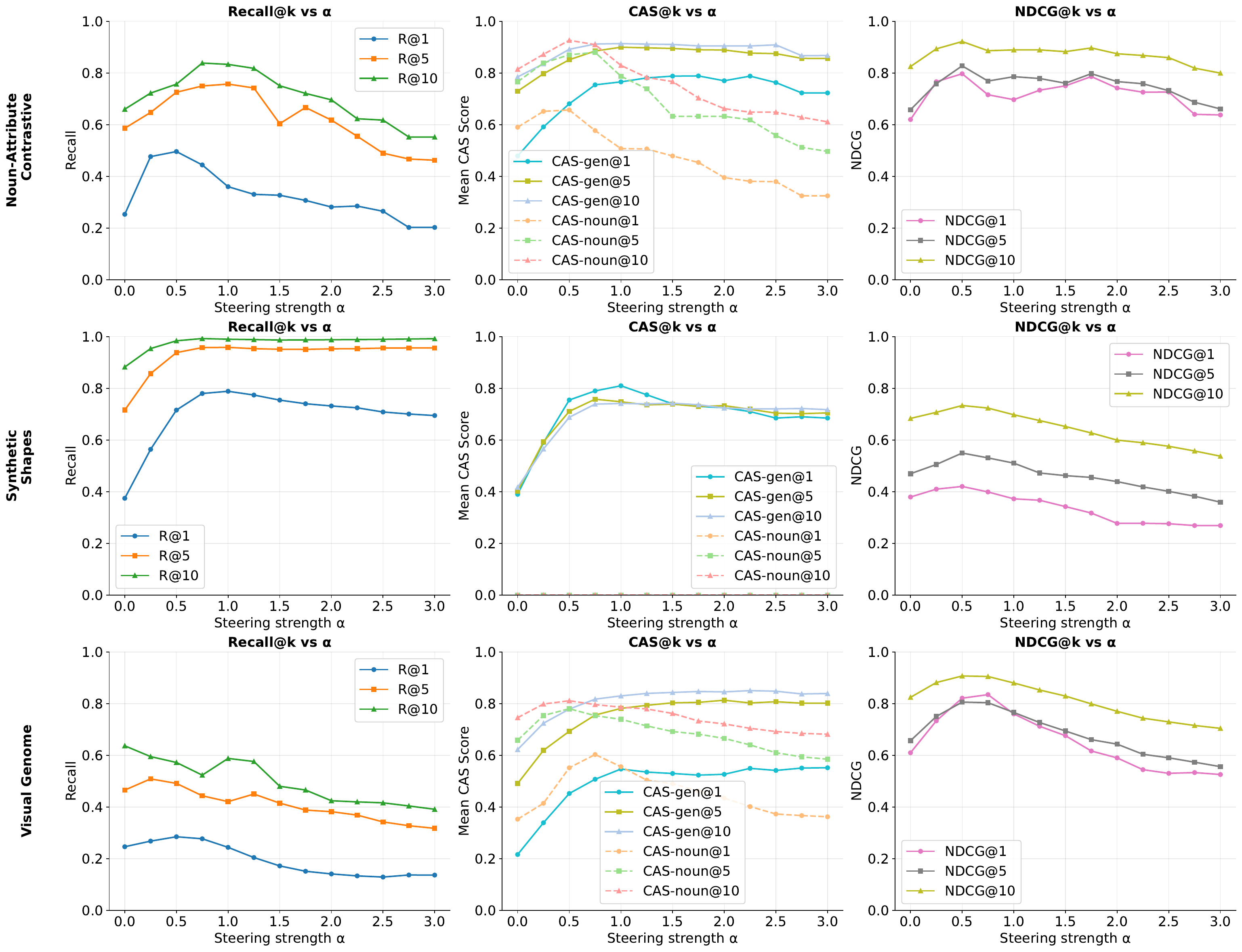}
    \caption{Retrieval quality as a function of steering strength $\alpha$ on NAC, Synthetic Shapes, and Visual Genome. For each dataset, we report Recall@K, concept attribution scores, and nDCG@K under query-conditioned steering. Steering is effective in single-concept settings, where moderate values of $\alpha$ improve both retrieval and concept fidelity, with especially strong gains on Synthetic Shapes and consistent improvements on NAC. In contrast, performance on Visual Genome degrades as $\alpha$ increases.}
    \label{fig:alpha_steer_metrics}
\end{figure}

Figure \ref{fig:fine_steering_sample} shows a sample from NAC dataset where steering strength $\alpha$ is varied to obtained finegrained retrieval. We vary $\alpha$ and capture the retrieval metrics on three datasets : NAC, Synthetic Shapes and Visual Genome. The results are captured in Figure \ref{fig:alpha_steer_metrics}.

\label{app:steering_concept_control}

\begin{figure}[h]  
    \centering
    \includegraphics[width=1\textwidth]{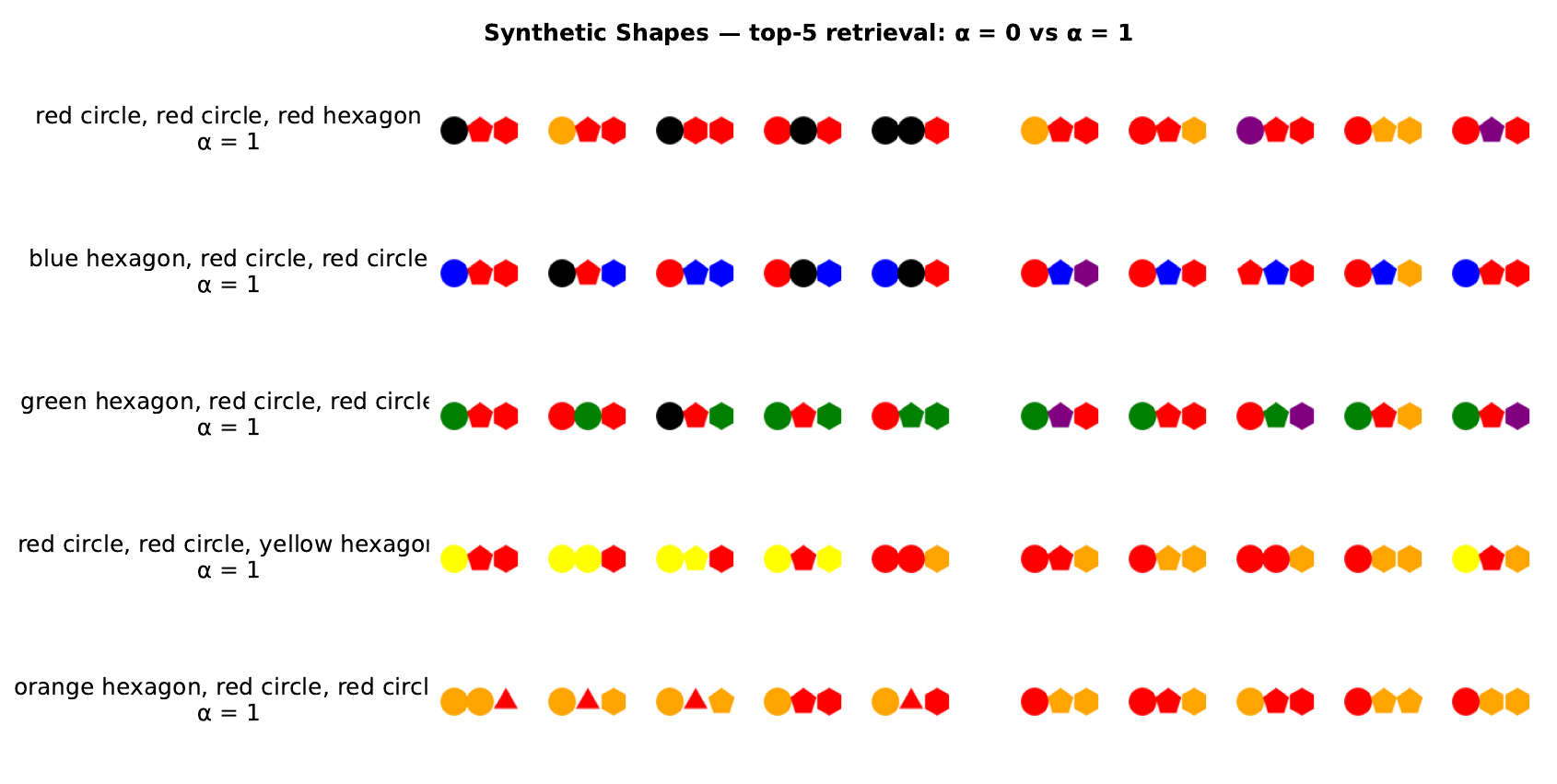}
    \caption{Sample Synthetic Shapes top-5 retrievals: original CLIP text--image matching ($\alpha = 0$, left) versus text embeddings steered toward the contrast attribute ($\alpha > 0$, right).}
    \label{fig:samples_shapes_steering}
\end{figure}

\begin{figure}[h]  
    \centering
    \includegraphics[width=1\textwidth]{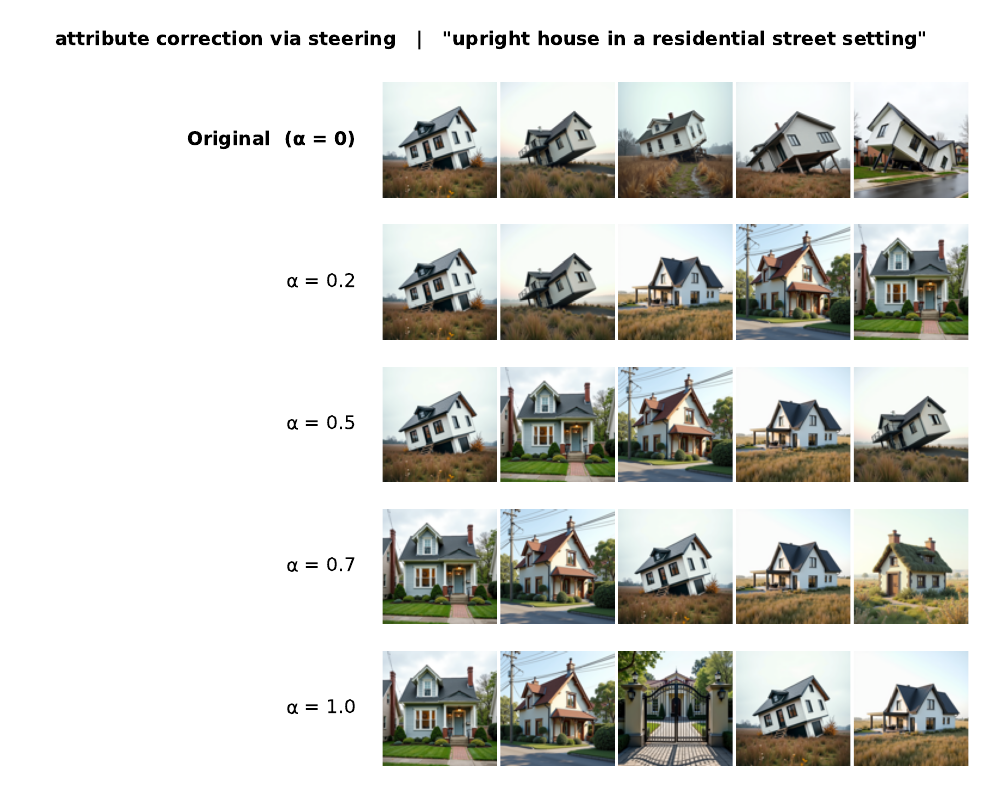}
    \caption{Sample NAC top-5 retrievals: original CLIP text-image matching ($\alpha = 0$) versus text embeddings steered to correct the attribute `upright' ($\alpha > 0$).}
    \label{fig:nac_steering_upright}
\end{figure}

\begin{figure}[h]  
    \centering
    \includegraphics[width=1\textwidth]{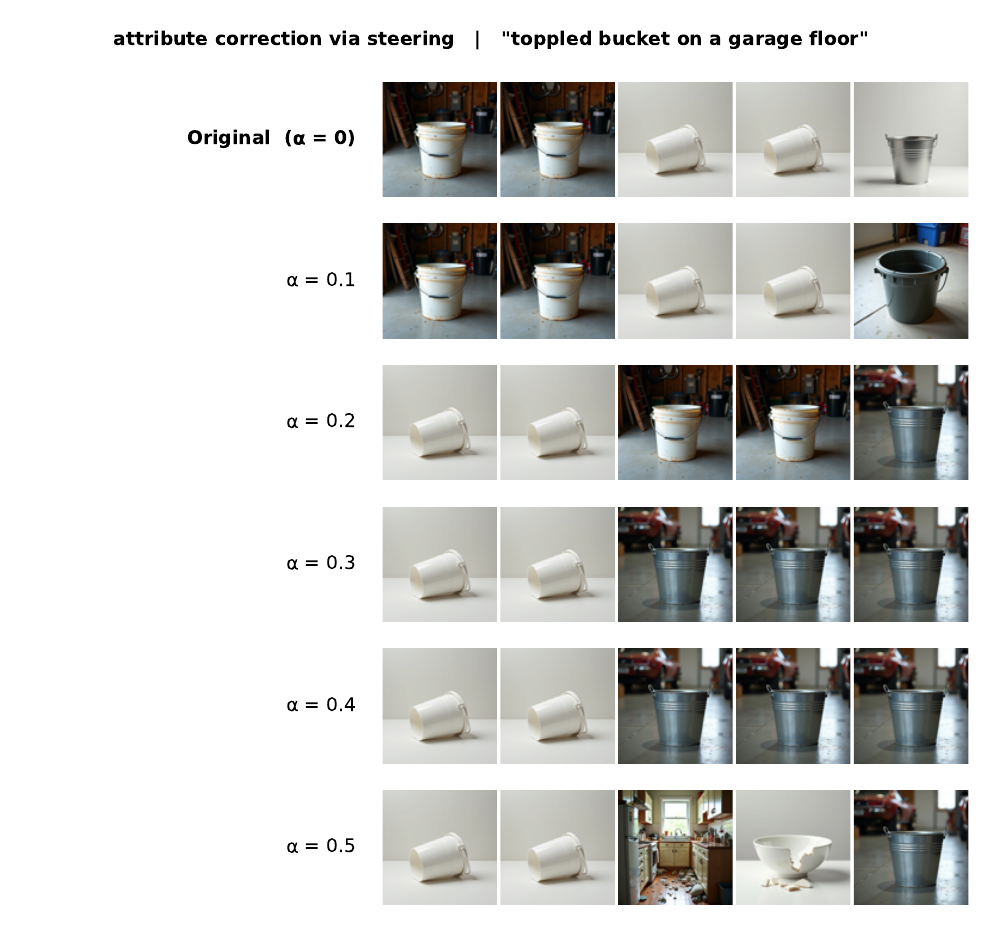}
    \caption{Sample NAC top-5 retrievals: original CLIP text-image matching ($\alpha = 0$) versus text embeddings steered to correct the attribute `toppled' ($\alpha > 0$).}
    \label{fig:nac_steering_toppled}
\end{figure}

\section{Compositional Steering Analysis}
\label{app:compositional_steering}

We study \emph{compositional image retrieval}: given a natural-language query describing multiple objects and their attributes, the goal is to retrieve the matching image from a database. The key challenge is not only recognizing individual attributes, but binding each attribute to the correct object. Standard vision--language retrieval models often fail on this binding problem, producing \emph{pairing errors}: retrieved scenes contain the correct attributes, but attached to the wrong objects.

\paragraph{Baseline CLIP retrieval.}
Plain cosine-similarity retrieval with holistic CLIP text and image embeddings yields a weak baseline of \textbf{retrieval@10 = 0.230}. Among the retrieved top-10 results, \textbf{10\%} are pairing errors: images that contain the right attributes but assign them to the wrong objects. Most retrieved results are complete misses: out of 10,000 retrieved images across the evaluation set, \textbf{9,644} do not match the target scene structure. Among pairing errors, color is the most frequently mispaired attribute (\textbf{42.2\%}), followed by shape (\(23.0\%\)), size (\(18.1\%\)), and material (\(16.7\%\)). This suggests that CLIP captures many relevant attribute values, but struggles to preserve object--attribute binding.

\paragraph{Targeted steering corrects local pairing errors.}
We test whether CLIP text space contains clean directions for correcting individual mispairings. For each mispaired object slot, we construct an attribute steering vector of the form
\[
\mathrm{encode}(\texttt{attr noun}) - \mathrm{encode}(\texttt{noun}),
\]
and apply it only to the affected object representation. On all \textbf{314 steerable cases} (color, material, and size), targeted steering achieves a \textbf{100\% correction rate}. This shows that CLIP text space contains semantically precise local attribute directions. However, this does \emph{not} translate into better full-query retrieval: query-level retrieval@10 drops to \textbf{0.144}. Shape errors (\textbf{94/408} mispaired slots) are not steerable in this formulation, since shape is entangled with noun identity rather than appearing as a separable attribute direction.

\paragraph{Naive multi-object steering fails.}
We next steer all objects in the query simultaneously and merge the resulting per-object representations into a single query vector. Across all tested steering strengths, this performs substantially worse than the CLIP baseline. With average merging, retrieval@10 ranges from \textbf{0.046} to \textbf{0.094}, with the best result at \(\alpha=0.7\), still more than \(2\times\) worse than plain CLIP retrieval. This indicates that although individual attribute directions are locally meaningful, naively aggregating multiple steered object vectors destroys compositional structure and drifts away from the image embedding space.

\paragraph{Two-stage reranking with min/softmin remains ineffective.}
To test whether steering can help after retrieval, we perform two-stage reranking. CLIP first retrieves a candidate pool of 100 images, achieving \textbf{candidate recall@100 = 0.669}. We then score each candidate using per-object steered query vectors, aggregated with either \texttt{min} or \texttt{softmin}. Despite the reasonably strong candidate pool, reranking remains ineffective: retrieval@10 drops to \textbf{0.089} with \texttt{min} and \textbf{0.092} with \texttt{softmin}. Thus, even when the correct image is often present among the candidates, simple per-object score aggregation does not produce a reliable compositional ranking signal.

\paragraph{FGW reranking resolves compositional structure.}
Our strongest result comes from replacing scalar aggregation with \emph{Fused Gromov--Wasserstein} (FGW) reranking. We first retrieve \(k=50\) candidates with CLIP, then compare the query and each candidate as \emph{sets of object representations}. FGW jointly aligns feature similarity between corresponding objects and the relational structure between objects, captured by pairwise distances within each set. This yields \textbf{retrieval@10 = 0.521}, more than doubling the CLIP baseline (\(0.230\)). Candidate recall@50 is \textbf{0.522}, nearly identical to final FGW retrieval@10, indicating that FGW is highly effective whenever the correct image appears in the candidate pool.

\paragraph{Interference analysis.}
To understand why multi-object steering fails, we analyze attribute-level interference under average-merged steering. The degradation is nearly universal: \textbf{998/1000} queries suffer degradation on at least one attribute, and among queries that show any improvement, an average of \textbf{10.93} other attribute slots degrade simultaneously. The effect grows with scene complexity: for scenes with at least 6 objects, \textbf{every query} experiences degradation. Color is the most disruptive attribute; when color improves, shape degrades in \textbf{81\%} of such cases and material degrades in \textbf{80\%}. The degradation frequency remains near \(1.0\) across all steering strengths, suggesting that the failure is structural rather than a matter of steering magnitude.

\paragraph{Noun-level Concept Alignment Score (CAS-noun)@K.}
CAS-noun measures whether the retrieved images contain the \emph{full query objects} as grounded scene entities, rather than merely containing the right attributes somewhere in the scene. Let
\[
\mathcal{O}(q) = \{o_1,\dots,o_m\}
\]
denote the query objects, where each object is represented as a full attribute tuple
\[
o = (\texttt{color}, \texttt{material}, \texttt{shape}, \texttt{size}).
\]
For a retrieved image \(x\), let
\[
\mathcal{T}(x) = \{t_1,\dots,t_n\}
\]
denote the corresponding set of object tuples extracted from its scene graph. We define
\[
\mathrm{CAS\mbox{-}noun}(x,q)
=
\frac{1}{|\mathcal{O}(q)|}
\sum_{o \in \mathcal{O}(q)}
\mathbbm{1}[\,o \in \mathcal{T}(x)\,].
\]
Aggregating over the top-\(K\) retrieved images gives
\[
\mathrm{CAS\mbox{-}noun@}K(q)
=
\frac{1}{K}
\sum_{i=1}^{K}
\mathrm{CAS\mbox{-}noun}(x_i, q),
\]
and averaging over all queries gives
\[
\mathrm{Mean\mbox{-}CAS\mbox{-}noun@}K
=
\frac{1}{|Q|}
\sum_{q \in Q}
\mathrm{CAS\mbox{-}noun@}K(q).
\]
CAS-noun is stricter than CAS-gen: CAS-gen can be high when the retrieved set contains the correct attributes in aggregate, even if they are attached to the wrong objects, whereas CAS-noun requires that the full attribute bundle of a query object appear on a single scene object.

\paragraph{Summary.}
These results reveal a clear hierarchy. Individual attribute steering directions in CLIP text space are highly precise and can correct local pairing errors, but they do not compose cleanly across multiple objects. Naive merging of per-object steering vectors introduces severe interference and harms retrieval, even when used only for reranking. FGW succeeds because it avoids collapsing the query into a single vector; instead, it compares query and image as structured sets and aligns them via optimal transport. This suggests that the main obstacle in compositional retrieval is not the absence of useful attribute directions, but the lack of a retrieval mechanism that preserves and compares \emph{object-level structure}.